\documentclass[journal]{IEEEtran}
\usepackage{lmodern}
\usepackage{enumitem}
\usepackage{graphicx}
\usepackage{mathtools}
\usepackage{float}
\usepackage{multirow}
\usepackage{geometry}
\usepackage{booktabs}
\usepackage{subfigure}
\usepackage{amsmath}
\newcommand{\tabincell}[2]{\begin{tabular}{@{}#1@{}}#2\end{tabular}}

\ifCLASSINFOpdf
\else
\fi
\begin{document}
%
\title{DWnet: Deep-Wide Network for 3D Action Recognition}
%
%
%

\author{Yonghao Dang,
        Fuxing Yang,
        and Jianqin Yin
\thanks{Y. Dang, Fuxing Yang and Jianqin Yin are with the Department of Automation, Beijing University of Posts and Telecommunication, Beijing, 100876, China.}
\thanks{Jianqin Yin is the corresponding author.}}
\maketitle

\begin{abstract}
We propose in this paper a deep-wide network (DWnet) which combines the deep structure with the broad learning system (BLS) to recognize actions. Compared with the deep structure, the novel model saves lots of testing time and almost achieves real-time testing. Furthermore, the DWnet can capture better features than broad learning system can. In terms of methodology, we use pruned hierarchical co-occurrence network (PruHCN) to learn local and global spatial-temporal features. To obtain sufficient global information, BLS is used to expand features extracted by PruHCN. Experiments on two common skeletal datasets demonstrate the advantage of the proposed model on testing time and the effectiveness of the novel model to recognize the action.
\end{abstract}


%
\IEEEpeerreviewmaketitle

\section{Introduction}
Human action recognition plays an important role in describing human gesture and predicting human behavior. Recently, researches about human action recognition have achieved wonderful results, which have been applied to computer vision, such as Intelligent monitoring, Abnormal Behavior Detection and so on \cite{Lee2017Ensemble, Tang_2018_CVPR}. Given the fast development of lowcost devices to capture 3-dimensional (3D) data (e.g. camera arrays and Kinect), increasingly more researches are actively conducted over 3D action recognition \cite{Ke2017,8026285,Si2018,Zhang2018,Zhu2016}. 3D skeleton not only provides more spatial information, but also is robust to variations of viewpoints, human body scales, and movement speed \cite{HAN201785}, so the 3D skeleton has attracted more and more attention in recent years \cite{Tang_2018_CVPR,Pei2016,8099538,6239233}. In this paper, we mainly focus on recognizing actions based on 3D skeleton.

 Recently, deep structures, such as Recurrent Neural Network (RNN) \cite{Graves2012,6638947}, Long Short-Term Memory (LSTM)\cite{7178838,Vivek2015}, Convolutional Neural Network (CNN) et al.\cite{7742919,Ke2017,8026285}, have been widely used to recognize actions via 3-dimensional (3D) skeleton. However, RNN is difficult to capture the long-term temporal information \cite{7486569,article}. And it is difficult for CNN to model the global spatial features\cite{Ke2017,IJCAI2018}. To mine co-occurrences from all joints efficiently, Li et al. \cite{IJCAI2018} designed the Hierarchical Co-occurrence Network (HCN) that is able to aggregate different levels of contextual information. The deep structure has the ability to extract multi-scale local spatial-temporal features and superior approximation ability due to its local connection, shared weights and the use of multi-layer. However, with the depth increasing, it is possible for deep structure to lose some information and spend more time to solve the optimal solution.

 Broad learning system (BLS) proposed by Chen and Liu \cite{7987745} is a kind of flat fully connected network. In BLS, matrix multiplication is the main operation. There is an activation function during generating enhancement nodes. The approximation ability of the model can be improved by adjusting the number of non-linear enhancement nodes. And feature nodes and enhancement nodes are connected to the output, which makes it easy to accumulate features.
\begin{figure}[htb]
  \centering
  \includegraphics[scale=0.5]{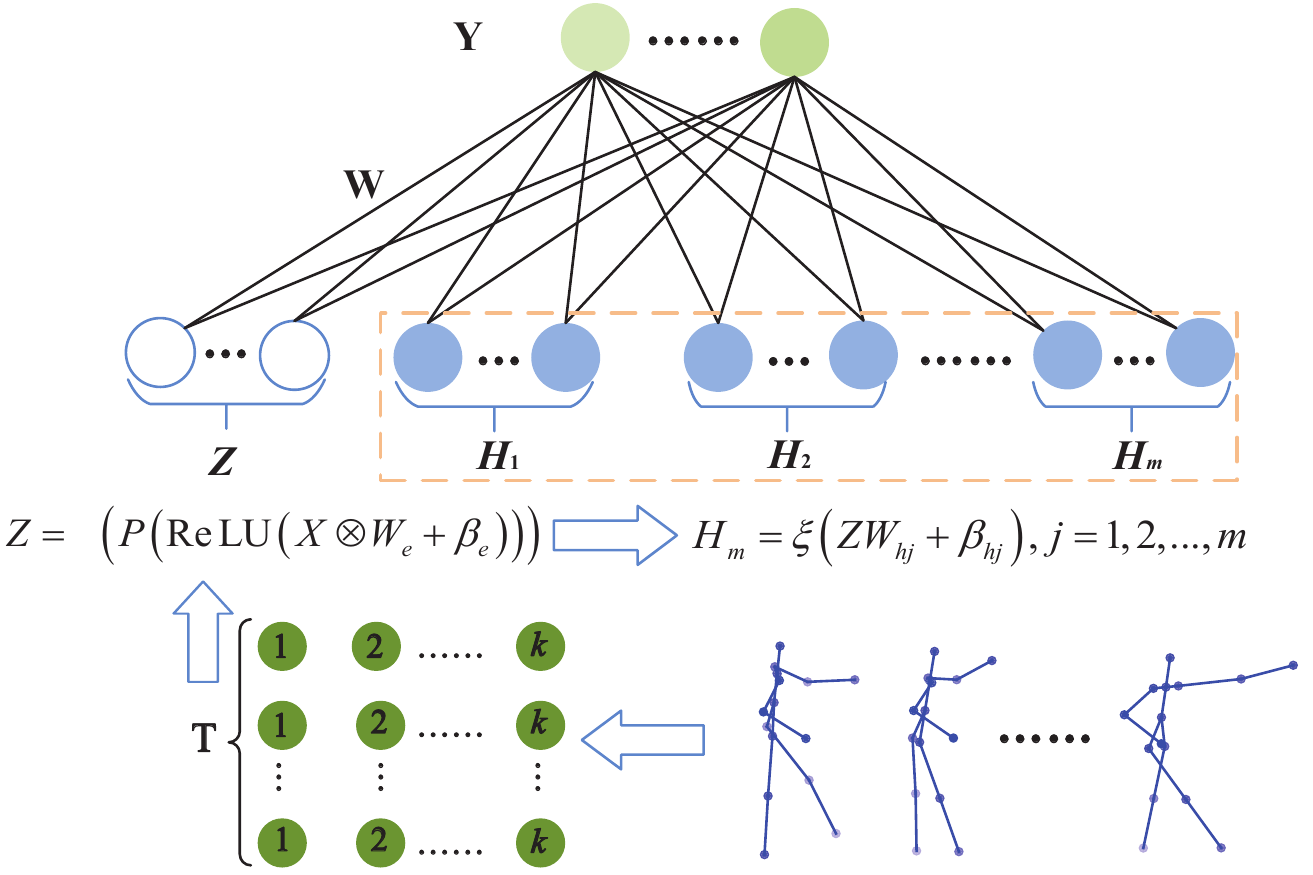}
  \caption{The overall pipeline of deep-wide network. Firstly, the coordinates of skeleton are processed into images with the size of $T \times k \times 3$, where the $T$ is the number of frames representing the height of the image, $k$ is the number of joints representing the width of the image, and 3 is the dimension of the coordinate representing the number of channels of the image. Then the pruned HCN (PruHCN) is used to extract spatial-temporal features and co-occurrence features of joints. Next the features learned by PruHCN are enhanced by enhancement nodes generated through random mapping. Finally, features extracted by PruHCN and enhancement nodes are connected to the output layer to recognize the action.}
  \label{BLS_CV_a}
\end{figure}
 The desired weights connected to the output are determined by a fast ridge regression of the pseudo-inverse of the system equation \cite{JIN2018}. The BLS is mainly aimed to approximate the training labels, and the underlying geometrical structure of data is not fully considered \cite{JIN2018}. Although the BLS is effective on benchmark datasets like MNIST and NORB \cite{8091501}, it is rarely used for recognizing actions.

 Considering that the HCN can extract effective spatial-temporal co-occurrence features from skeletal data, the BLS is efficient to aggregate features, we propose a new model named deep-wide network (DWnet) by combining these two structures, as it is shown in Fig. \ref{BLS_CV_a}. The deep structure adopts pruned HCN (PruHCN) that is produced by pruning  layers of the HCN to extract skeletal features. And then the PruHCN is used as the mapping function to generate the feature nodes of the BLS. Next, the feature nodes are expanded to the enhancement nodes to reinforce the model. Finally, all the feature nodes and enhancement nodes are connected to the output to recognize actions. The proposed model can keep the approximation ability utilizing non-linear enhancement nodes provided by BLS. Furthermore, the DWnet is more time-saving than the deep structure.Contributions in this paper are as follows:
\begin{itemize}
 \item We propose a new structure, DWnet, which combines deep structures with broad learning system to recognize actions based on 3D skeletal data.
 \item The presented model greatly saves testing time and achieved comparable results on SBU Kinect dataset and NTU RGB+D dataset.
\end{itemize}

\section{Related Work}
In this section, we will review the related work from two aspects: one is action recognition based on deep learning and 3D skeleton, the other is the application of broad learning system (BLS) in image processing.

\subsection{Action Recognition Based on deep learning Using 3D Skeleton}

There have been lots of deep learning researches based on 3D skeleton data, which can be roughly divided into two categories: one is to utilize recurrent neural network for action recognition and the other is to apply convolutional neural network for action recognition.

For example, Shahroudy and Liu et al. \cite{NTU_RGB} designed a new recurrent neural network to capture the long-term temporal features on RGB+D dataset, and obtained satisfied results. Liu and Shahroudy et al. \cite{Liu2016} extended RNN to spatial-temporal domains to catch static features and motion features respectively, in addition, associating gating mechanism with Long Short-Term Memory networks (LSTMs) has promoted the robustness to noise and occlusion on skeleton data. The fully connected deep LSTM with a novel regularization scheme was introduced to learn co-occurrence feature of different joints in \cite{IJCAI2018}.

Recurrent neural networks can extract temporal features, but it is difficult to model high-level spatial features. In the past few years, convolutional neural networks (CNNs) are widely applied to recognize actions based 3D skeleton data. Hou and Li et al. \cite{7742919} encoded the skeletal sequence into color texture images, and then employed CNN to catch the discriminative features for action recognition. In \cite{LIU201693}, Liu and Zhang et al. constructed a 3D-based deep convolutional neural network to gain spatial-temporal features from the original sequence, and then calculated joint vector as the input of support vector machine for action recognition. Recently, Ke and Bennamoun et al. \cite{Ke2017} transformed raw skeleton sequence to three clips that are fed to a deep CNN to extract long-term features. Nevertheless, splitting sequence into several clips will drop some global information. In \cite{8026285}, Li and Zhong et al. designed a novel 7-layer CNN for action recognition and detection. And then Li and Zhong et al. \cite{IJCAI2018} proposed Hierarchical Co-occurrence Network (HCN) to learn the co-occurrence features that is contextual information in different levels. At present, HCN has outperformed others among the methods mentioned above. And the HCN has reached the state-of-the-art results on both action recognition and detection.

\subsection{Applications of BLS in Image Processing}

Chen and Liu \cite{7987745} proposed an efficient and effective discriminative learning method called Broad Learning System (BLS). BLS was designed inspired by random vector functional-link neural network (RVFLNN) \cite{RVFLNN}. And it has shown superior performance on MNIST and NORB dataset, indicating that BLS has fantastic approximation ability. Furthermore, in \cite{8457525}, Chen and Liu et al. proposed several variants of broad learning system, that makes it flexible to apply BLS to many fields, especially, the BLS has been widely used to image processing in the past few years. Liu and Zhou et al. \cite{8091501} employed a novel broad learning structure based on the K-means to cluster images in CIFAR-10 dataset. Dang and Wang et al. \cite{8616713} applied broad learning system to estimate cement compressive strength from microstructure images of cement. In \cite{8520880}, BLS was utilized for estimating sun visibility from given outdoor images, and then incremental broad learning system, the improvement of broad learning system proposed in \cite{7987745}, was applied to classify the sun visibility into more categories. Considering BLS is able to model the large-scale data, Shi and Wei et al. \cite{Shi2019} designed fisher broad learning system (FBLS), and adopted Local Log-Euclidean Multivariate Gaussian (L2EMG) \cite{7463054} for student gesture recognition task. A novel variant graph regularized broad learning system (GBLS) was introduced by Jin and Liu et al. \cite{JIN2018} for face recognition, in which the manifold technology is incorporated into the optimization process of BLS.

 The HCN has the ability to extract invariant, complex co-occurrence features. And the HCN improves the approximation ability by increasing layers. The BLS is able to keep the approximation ability of the model by increasing the number of enhancement nodes. Furthermore, the BLS is more efficient than the deep structure. Thus, we take pruned HCN (PruHCN) as the mapping function of the BLS to combine the deep structure with the width structure in this paper. And details of the presented method will be discussed in next section.

\begin{figure}[htb]
  \centering
  \includegraphics[scale=0.5]{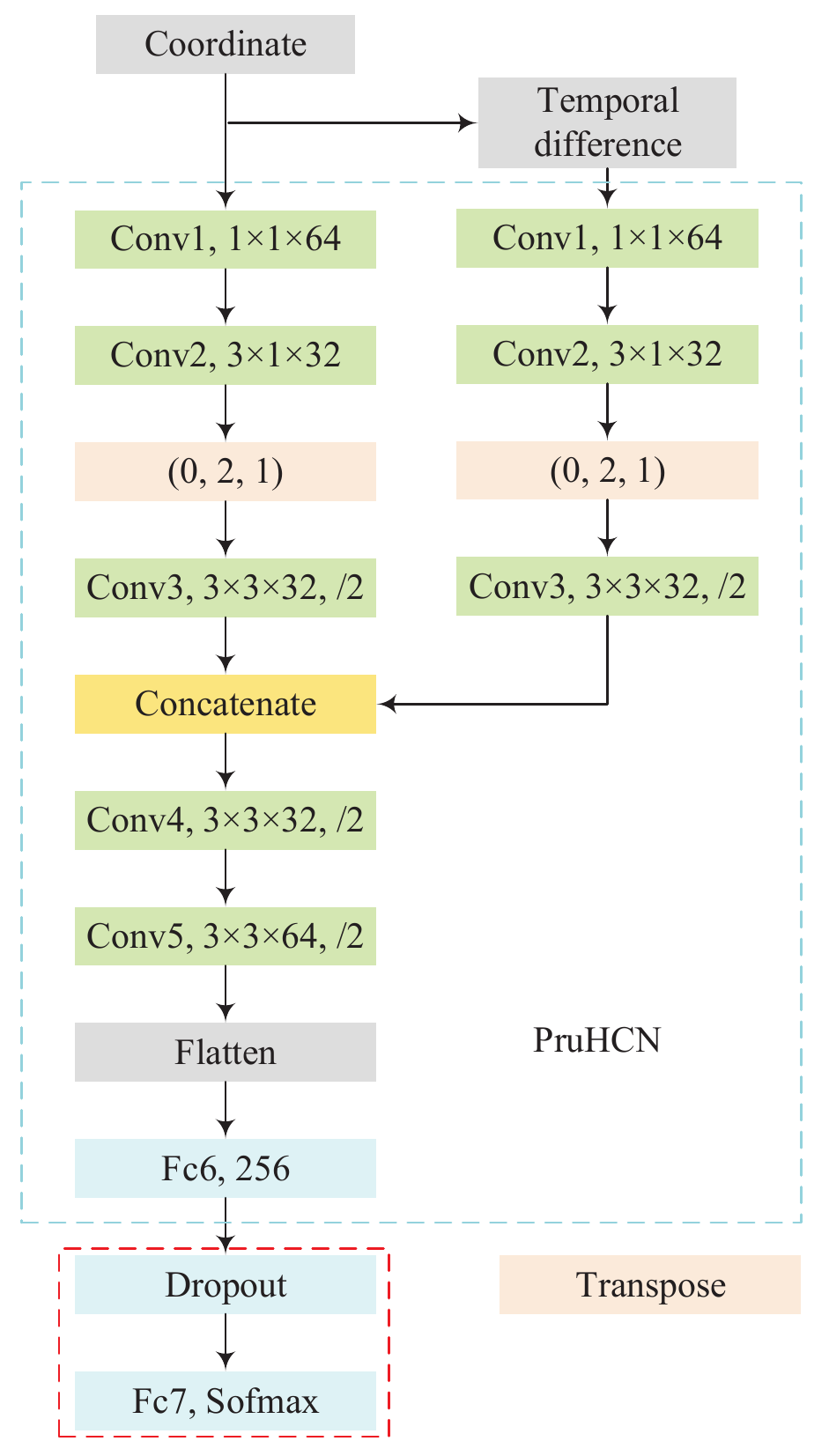}
  \caption{The structure of HCN and PruHCN. The structure in the blue box is the PruHCN. Compared with the HCN, PruHCN prune the Dropout layer and the ``Fc7'' layer which in the red box. The ``/2'' represents the MaxPooling layer with stride 2.}
  \label{BLS_CV_b}
\end{figure}

\section{Methodology}
 The PruHCN is used as the feature mapping function to extract features of the preprocessed skeleton in the proposed model. And then features learned by PruHCN are enhanced to enhancement nodes. All the feature nodes and enhancement nodes are connected to the output to recognize the action.

In this section, we first describe the data preprocessing, and then introduce the structure of pruned hierarchical co-occurrence network. Finally, the implementation of our model is described in detail.

\subsection{Data Preprocessing}
 The structure of the skeletal data is irregualr, while the CNN can only effectively deal with the regular Euclidean data like the image and the text \cite{Zhou2018Graph}. Before entering the model, the skeletal data is encoded into the image. Firstly, the skeletal points of each frame are stretched into a vector. Secondly, the image is obtained by concatenating $T$ frames into one matrix. The size of the encoded image is fixed to $T \times K \times c$, where $T$ represents the number of frames in one image, $K$ is the number of joints, and $c$ is the dimension of the coordinates. The high, width and channels of the image are represented by the $T, K$ and $c$ respectively. In this way, the spatial coordinates are converted to the information of channels of images \cite{IJCAI2018}.

 As for the temporal feature, the motion information is represented by the positional change of the corresponding skeletal point between adjacent frames. We use $M^{i}$ to describe the motion information at the $i$th moment. And the motion information is described as the temporal difference of each joint between contiguous frames \cite{IJCAI2018}, and it can be defined as follows:

\begin{equation}
    \begin{split}
        M^{i} &= {S^{i + 1}} - {S^{i}} \\
        &= \left\{{J_{1}^{i + 1} - J_{1}^{i},J_{2}^{i + 1} - J_{2}^{i},..., J_{K}^{i + 1} - J_{K}^{i}}\right\}
    \end{split}
    \label{eq:motion}
\end{equation}

 where $S_{i} = {J_{1}^{i}, J_{2}^{i},\cdots, J_{K}^{i}}$, in which $i$ represents the frame at the $i$th moment, $K$ is the number of joint, and $J_{K}^{i}=\left(x, y, z\right)$ is the spatial coordinate of the joint \cite{IJCAI2018}.

\subsection{Pruned hierarchical co-occurrence network}

 As shown in Fig. \ref{BLS_CV_b}, in original HCN model, there are seven layers including five convolutional layers and two fully connected layers. And the softmax function is used as the classifier in the last fully connected layer. The symbol ``/2'' denotes the pooling with step size 2. And the ReLU function is used as the activation function.

 There are many ways to prune layers of the HCN. Regulations of pruning are that firstly the pruned model has the ability to extract effective co-occurrence features; secondly, the complexity of the features learned by the model should be kept as low as possible.

 The pruned layers of the HCN include a Dropout layer and a fully connected layer. Since the convolutional operation has the characteristics of local connection and shared weights, each feature map obtained by convolutional operation contains local spatial-temporal information. Convolutional layers are retained in the PruHCN. The layers (``Conv1'', ``Conv2'') are used to extract the features of single joint. The convolutional layer (``Conv3'') is able to model the relationship between adjacent joints. The co-occurrence features are captured by the convolutional layer 4 and 5. The co-occurrence features extracted by the convolutional layer ``Conv5'' include sufficient local sapatial-temporal information.

 In order to integrate local features and reduce the dimension of features, the layer ``Fc6''  is used to aggregate local information contained in each feature map. The features processed by fully connected layer contain both local spatial-temporal features and global co-occurrence features of joints. Furthermore, after the layer ``Fc6'' processing, the dimension of the feature matrix is reduced, which simplifies the problem.

\subsection{Deep-Wide Network}
 The PruHCN inherits the advantages of HCN and can learn the co-occurrence characteristics of skeletal sequences. Although the PruHCN is two layers less than the HCN, the BLS can restore the approximation ability of the model in width. There are many ways to connect the PruHCN and the BLS, such as several PruHCNs are connected in parallel or series. The parallel method will cause the redundancy of extracted features and increase of noise, which will lead to the degradation of the performance to recognize the action. By connecting a trained PruHCN to the BLS in series, not only effective co-occurrence features can be extracted, but also the approximation ability of the model can be guaranteed by BLS.

 Therefore, the trained PruHCN is used as the feature mapping function to extract local and global spatial-temporal features. And then, the feature nodes are expanded to enhancement nodes to reinforce the model. Finally, all the feature nodes and enhancement nodes are connected to the output to recognize the action. Next, the model proposed in this paper is described in detail in terms of feature mapping and computational complexity.

\textbf{Feature Mapping}\quad As illustrated in Fig. \ref{BLS_CV_a}, the number of feature nodes is set to one. Input data is transformed to feature nodes by PruHCN at first. And the feature mapping function $\Phi\left(\cdot\right)$ can be defined as follows:

\begin{equation}
    \begin{split}
        {Z_{i}} &= \phi \left( {X;\left\{ {{W_{ei}},{\beta _{ei}}} \right\}} \right) \\
        &= \theta \left({P\left({\rm ReLU}\left({X \otimes {W_{ei}} + {\beta _{ei}}}\right) \right)} \right), \\
        & i = 1,2,...,n
    \end{split}
    \label{eq:conv}
\end{equation}

 where $\Phi\left(\cdot\right)$ is the mapping function, $X$ is the input data, $W_{ei}$ and $\beta_{ei}$ are weights and biases learned through training, $\otimes$ is the convolutional function for the given matrices, function $P\left(\cdot\right)$ is the pooling operation and $\theta\left(\cdot\right)$ is the selected activation function. In DWnet, we set $i=1$, i.e. we apply one trained PruHCN module to extract the spatial-temporal characteristics of the skeleton.

  Compared with the original HCN, due to cutting off the last two layers, the approximation ability of the PruHCN is not  as good as that of HCN. The BLS can improve the approximation ability of the model by increasing the number of enhancement nodes. In order to keep the approximation ability of the model and obtain discriminative features, the feature node is expanded to enhancement nodes. The mapping function $\xi\left(\cdot\right)$ can be defined as follows:

\begin{equation}
    \begin{split}
        {H_{j}} = \xi \left( {Z{W_{hj}} + {\beta _{hj}}} \right),j = 1,2,...,m
    \end{split}
    \label{eq:enhance}
\end{equation}

 where $Z$ is the feature node produced by PruHCN, $W_{hj}$ and $\beta_{hj}$ are the weights and biases generated randomly and $m$ is the number of enhancement nodes. The features extracted by PruHCN are expanded to higher dimension through converting into enhancement nodes, which is beneficial to recognize the action. Finally, the feature node and enhancement nodes are connected to the output. The flat fully connected structure makes it easy for DWnet to aggregate the global information.

 \textbf{Analysis of Computational Complexity}\quad In the DWnet, pruned layers include a Dropout layer and a Dense layer. In the Dropout layer, operations involve generating the vector that obeys Bernoulli distributions, linear operations on matrix and rescaling the result, with approximate computational complexity of $O\left(n\right)$, $O\left(n^{3}\right)$ and $O\left(n\right)$, respectively. The Dense layer contains linear operations on matrix, normalizing data, e-based exponential operation and classifying, with approximate computational complexity of $O\left(n^{3}\right)$, $O\left(n\right)$, $O\left(2^{n}\right)$ and $O\left(n\right)$. However, the BLS only includes $tansig$ activation function and linear operations on matrix, with approximate computational complexity of $O\left(2^{n}\right)$ and $O\left(n^{3}\right)$ ,respectively. The computational complexity of the BLS is lower than that of pruned layer of HCN. Therefore, replacing layers removed from the HCN with BLS can improve the efficiency of the model.

 DWnet can not only extract local spatial-temporal features effectively, but also be able to aggregate co-occurrence features over local aggregation efficiently. Compared with the HCN, although the DWnet prunes the Dropout layer and the ``Fc7'' layer, the BLS keeps the approximation ability of DWnet by increasing the number of non-linear enhancement nodes. Compared with the broad learning system, there is less randomness in DWnet during generating feature nodes, which ensures that the features learned by the DWnet are discriminative and representative.

 \section{Experiments}

 We evaluate the proposed framework on SBU Kinect Interaction \cite{6239234} dataset and NTU RGB+D dataset \cite{NTU_RGB}. In this section, we simply introduce datasets used in this paper. Next, baselines used in this paper are introduced briefly. And then we demonstrate the performance of the model proposed this paper from accuracy and efficiency of testing separately. Finally, we analyze the impact of the number of enhancement nodes on the model.

\subsection{Dataset and Experimental Settings}
\textbf{SBU Kinect Interaction Dataset}\quad SBU Kinect Interaction Dataset is the dataset created by Yun and Honorio et al. \cite{6239234} using Microsoft Kinect for two-person interactions. There are eight categories including seven participants and 21 groups of interactive actions. In addition, it contains 282 skeletal sequences and 6822 frames. And each frame includes 15 joints. We divide the data into five groups according to the method in \cite{6239234}, and evaluate proposed model by 5-fold cross validation.

\textbf{NTU RGB$+$D Dataset} \quad NTU RGB$+$D dataset is a large scale dataset for human 3D human activity analysis \cite{NTURGBD}. It contains 56880 samples including 60 classes. In NTU RGB$+$D, there are 25 skeletal points extracted from one subject. According to \cite{NTURGBD}, there are two recommended approached to evaluate the model, i.e. Cross-Subject (CS) and Cross-View (CV) \cite{IJCAI2018}. In the cross-subject setting, there are 40 participants which are divided into two groups on average to train and validate the model respectively. In the cross-view setting, samples are divided into 3 groups according to camera views. And there are 2 groups used to train and another group is used to test.

\textbf{Experimental Settings}\quad In this paper, experiments are tested on a Linux server that equips with Intel-i7 3.5GHz CPU, two Nvidia GeForce GTX 1080Ti graphic cards and 32 GB memory. Under the same experimental conditions, we carry out a series of comparative experiments.

On SBU Kinect dataset, we encode skeletal sequences into the image with a size of $16\times15\times3$ and divide the data into five groups according to the method in \cite{6239234}, and evaluate proposed model by 5-fold cross validation. There are 6 layers including 4 convolutional layers and 2 fully connected layers in the PruHCN. The dimension of features and the size of kernels are shown in Fig. \ref{BLS_CV_b}. The $\rm ReLU$ activation function is appended after $Conv1, Conv4, Conv5$ and $Fc6$ to introduce non-linearity \cite{IJCAI2018}. And the dimension of the feature processed by trained PruHCN is 64. The $tansig$ function is used as the activation function of enhancement nodes to introduce non-linearity.

On NTU RGB$+$D dataset, firstly, the sub-sequence of skeleton is cropped randomly with the ratio of 0.9. And then, the skeletal data is processed into the image with the size of $32\times25\times3$, i.e. the skeletal sequence is fix to 32 frames per image. And parameters of the PruHCN are kept the same settings in \cite{IJCAI2018}. The dimension of the feature processed by trained PruHCN is 256. The performance of proposed model is evaluated on cross-subject and cross-view respectively. The $tansig$ is chosen as activation function to introduce non-linearity during generating enhancement nodes.

\subsection{Introduction of Baseline}

\textbf{Hierarchical Co-occurrence Network}\quad The original hierarchical co-occurrence network (HCN) proposed by Li and Zhong et al. \cite{IJCAI2018} is used as the baseline in our experiment. For the sake of fairness, the structure and parameters of the HCN adopted in this paper are consistent with those of the original HCN in \cite{IJCAI2018} on the SBU Kinect dataset and the NTU RGB+D dataset, respectively.

\textbf{Broad Learning System}\quad The general broad learning system (BLS) is used as baseline to recognize actions. On SBU Kinect dataset, the encoded image with the size of $16 \times 15 \times 3$ is expanded to an vector with the size of $1 \times 720$ containing one person, and the corresponding motion information is also processed into a vector with the size of $1 \times 720$. The size of final input is $N \times 2880$ including the spatial characteristics and motion information of two people, where $N$ is the number of samples. The number of feature nodes is 100, and the number of enhancement nodes is set to 8000. The $tansig$ activation function is used to introduce non-linearity.

\begin{figure}[htb]
  \centering
  \includegraphics[scale=0.4]{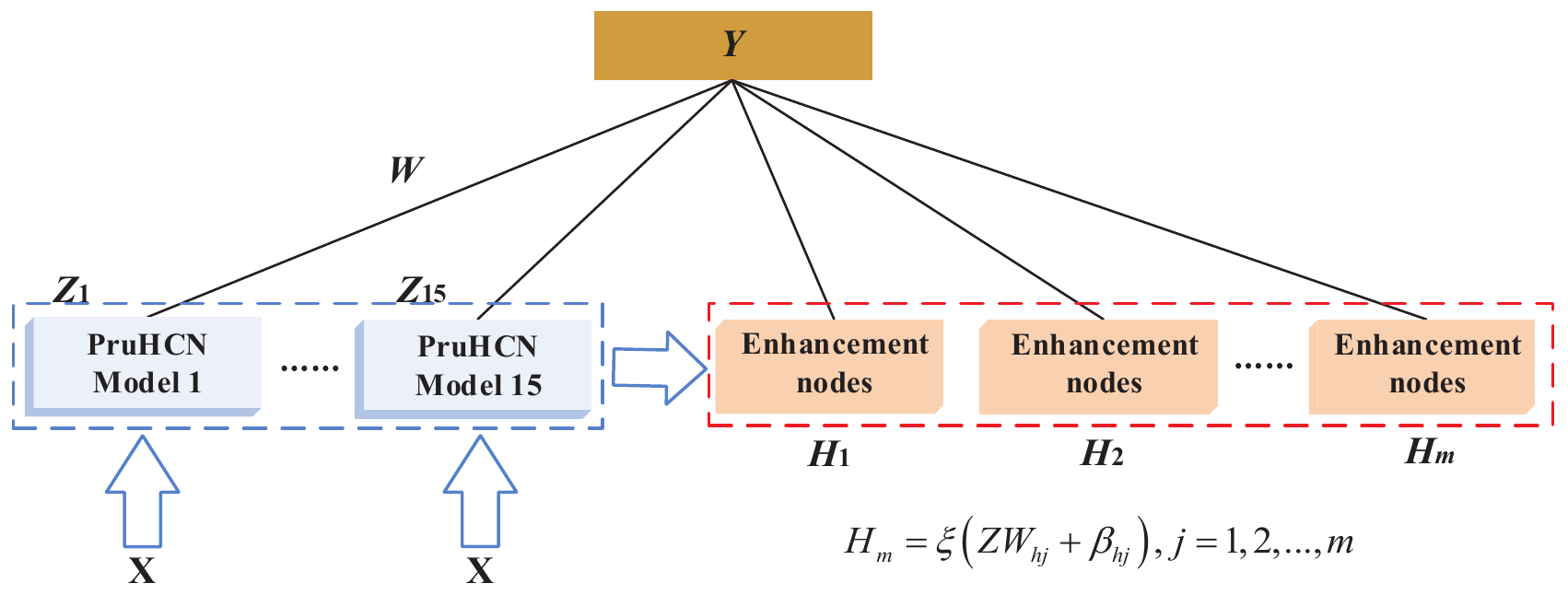}
  \caption{The structure of HCNBLS. It is the other structure that we explored which incorporates the deep structure and the wide structure}
  \label{fig:HCNBLS}
\end{figure}

\textbf{BLS with HCN}\quad In this paper, we aim to explore the method of combining the deep structure with the wide structure. And as shown in Fig. \ref{fig:HCNBLS}, it is the other structure that we explored which incorporates the deep structure and the wide structure, and we name it as  HCNBLS. The HCNBLS and DWnet are two different structures we proposed to combine deep and wide structures. The HCNBLS uses untrained PruHCN. And weights and biases are generated randomly in HCNBLS. The training algorithm adopts the training strategy described in BLS \cite{7987745}. However, the features extracted from only one untrained PruHCN may not be the most representative features. It is arduous to find out the optimal solution. Therefore, the recognizing accuracy is low if only one untrained PruHCN is used. To improve the representativeness of mapping features, 15 PruHCN models with separate inputs are used. The dimension of input of each PruHCN model is the same as the HCN. And the number of enhancement nodes is set to 550.

\subsection{Evaluation on SBU Kinect Interaction dataset}

\textbf{Evaluation on Accuracy}\quad We compare the performance of the DWnet to the previous HCN, BLS and HCNBLS. TABLE \ref{tab:table1} has shown the recognizing accuracy of each model on SBU Kinect Interaction dataset. Compared with the HCN, the average accuracy of the DWnet is improved by 2.56\%, which means that combining the deep structure with the BLS can keep and even improve the approximation ability of the raw deep model. In addition, compared with the DWnet, the performance of HCNBLS degrades about 10 percent. At the same time, the performance of the BLS degrades a lot. These results show that it is difficult for the BLS to extract discriminative features directly from the irregular skeletal data, even if the skeletal data is encoded into images. Thus, the results of the BLS and the HCNBLS are greatly worse than that of the DWnet.

 Compared with the BLS, the performance of HCNBLS increases about 20 percent. The reason is that the convolutional operation is chosen as the feature mapping function, which is more beneficial to extract effective features. However, the recognizing accuracy of HCNBLS is about 7 percent and 10 percent lower than that of HCN and DWnet, respectively. The possible reason for the decline is that the HCNBLS adopts the training strategy introduced in \cite{7987745}, and it is difficult for this training strategy to obtain the optimal solution at a deep level. But, the PruHCN of DWnet and the HCN are trained by using the stochastic gradient descent algorithm to find an optimal solution. Therefore, DWnet and HCN perform better.

\begin{table}
  \renewcommand\arraystretch{1.7}
  \caption{Testing accuracy on the SBU Kinect dataset.}
  \label{tab:table1}
  \begin{tabular}{cccccccc}
    \toprule
    Data & HCN & BLS & HCNBLS & DWnet \\
    \midrule
     $Test_1$ & 94.02\% & 62.69\% & 94.03\% & \textbf{97.02\%} \\
     $Test_2$ & 94.23\% & 73.08\% & \textbf{96.15\%} & 94.23\% \\
     $Test_3$ & 94.12\% & 64.71\% & 89.71\% & \textbf{98.53\%} \\
     $Test_4$ & 96.97\% & 66.67\% & 81.82\% & \textbf{98.49\%} \\
     $Test_5$ & 94.81\% & 72.73\% & 87.01\% & \textbf{98.70\%} \\
     $Ave$    & 94.83\% & 67.98\% & 87.944\% & \textbf{97.39\%} \\
    \bottomrule
  \end{tabular}
\end{table}

\begin{figure}[htb]
    \centering
    \includegraphics[scale=0.5]{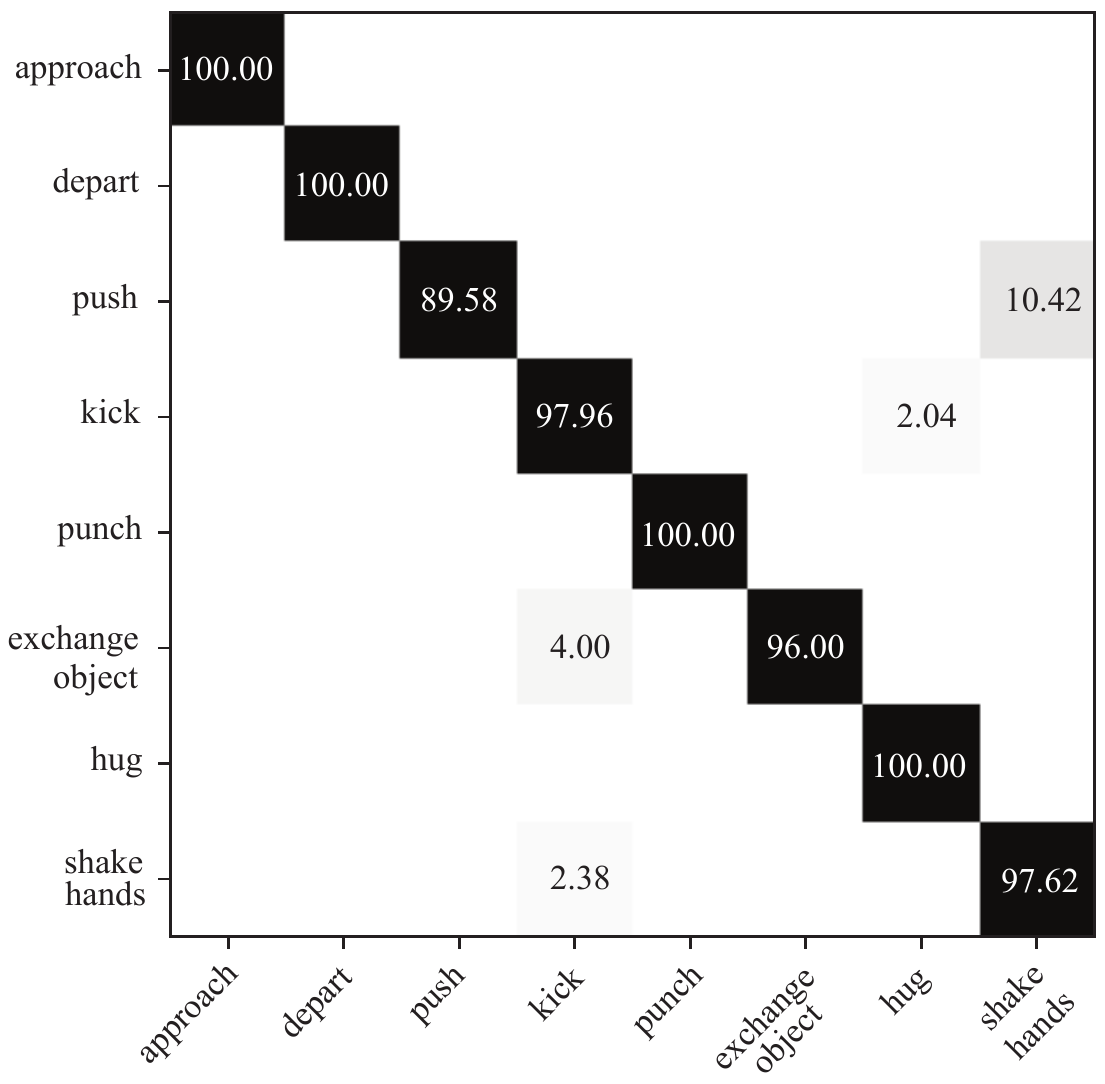}
    \caption{The average confusion matrix. The horizontal axis represents the predicted results of DWnet, and the vertical axis represents the actual label of actions. Diagonal elements indicate the accuracy of each type of action being recognized. }
    \label{fig:confusion}
\end{figure}

In addition, the confusion matrix of the DWnet is shown in Fig. \ref{fig:confusion}. The horizontal axis represents the predicted results of DWnet, and the vertical axis represents the actual label of actions. Therefore, diagonal elements indicate the accuracy of each type of action being recognized. It can be seen from Fig. \ref{fig:confusion} that our approach works very well for most of the actions. But there are 4 kinds of actions are easily confused. The ``push'' is recognized as ``shake hands''. The ``kick'' is classified as ``hug''. The ``exchange object'' and ``shake hands'' are recognized ``kick''. It is worth noticing that there is a similarity between actions , such as ``shake hands'' and ``push'', or the skeleton tracker fails frequently, such as the action ``kick'' and ``exchange object'', which makes it possible for DWnet to recognize the ``exchange object'' as other categories.

\textbf{Evaluation on Efficiency}\quad We compare the efficiency of each model under the same experimental environment. To be fair, we record the average testing time for each sample on the dataset. TABLE \ref{tab:timetable1} shows the testing time of each model. It can be seen from the TABLE \ref{tab:timetable1} that the efficiency of DWnet is about 100 times higher than that of HCN. As we discussed in section 3.3, compared with the HCN, the DWnet takes place the last two fully connected layers with the BLS. The computational complexity of the BLS is lower than that of the last two layers of HCN, so the DWnet is more efficient than the HCN. In addition, the HCNBLS consumes the most time to test. Compared with the HCNBLS, the DWnet only uses one PruHCN module to extract features, while the HCNBLS needs 15 even more PruHCN modules to capture sufficient features, so the dimension of the features extracted by the DWnet is much lower than that extracted by the HCNBLS. The higher dimensional feature matrix makes HCNBLS need more time to calculate the optimal solution. Therefore, compared with the DWnet and the HCN, HCNBLS takes more time to test. And TABLE \ref{tab:timetable1} demonstrates that the DWnet is efficient during the testing stage and can almost achieve real-time testing on SBU Kinect Interaction dataset.

\begin{table}
  \renewcommand\arraystretch{1.7}
  \caption{Testing Time on the SBU Kinect dataset. The unit of time is seconds/one sample.}
  \label{tab:timetable1}
  \begin{tabular}{cccccccc}
    \toprule
     Data & HCN & BLS & HCNBLS & DWnet \\
    \midrule
     $Test_1$ & 4.30e-2 & 8.96e-5 & 2.41 & 2.09e-4 \\
     $Test_2$ & 4.99e-2 & 1.15e-4 & 3.07 & 2.31e-4 \\
     $Test_3$ & 3.86e-2 & 5.88e-5 & 2.38 & 2.21e-4 \\
     $Test_4$ & 3.98e-2 & 6.06e-5 & 2.44 & 2.42e-4 \\
     $Test_5$ & 3.46e-2 & 9.09e-5 & 2.10 & 1.82e-4 \\
     $Ave$    & 4.12e-2 & 8.30e-5 & 2.48 & 2.17e-4 \\
    \bottomrule
  \end{tabular}
\end{table}

\begin{figure}[htb]
  \centering
  \includegraphics[scale=0.4]{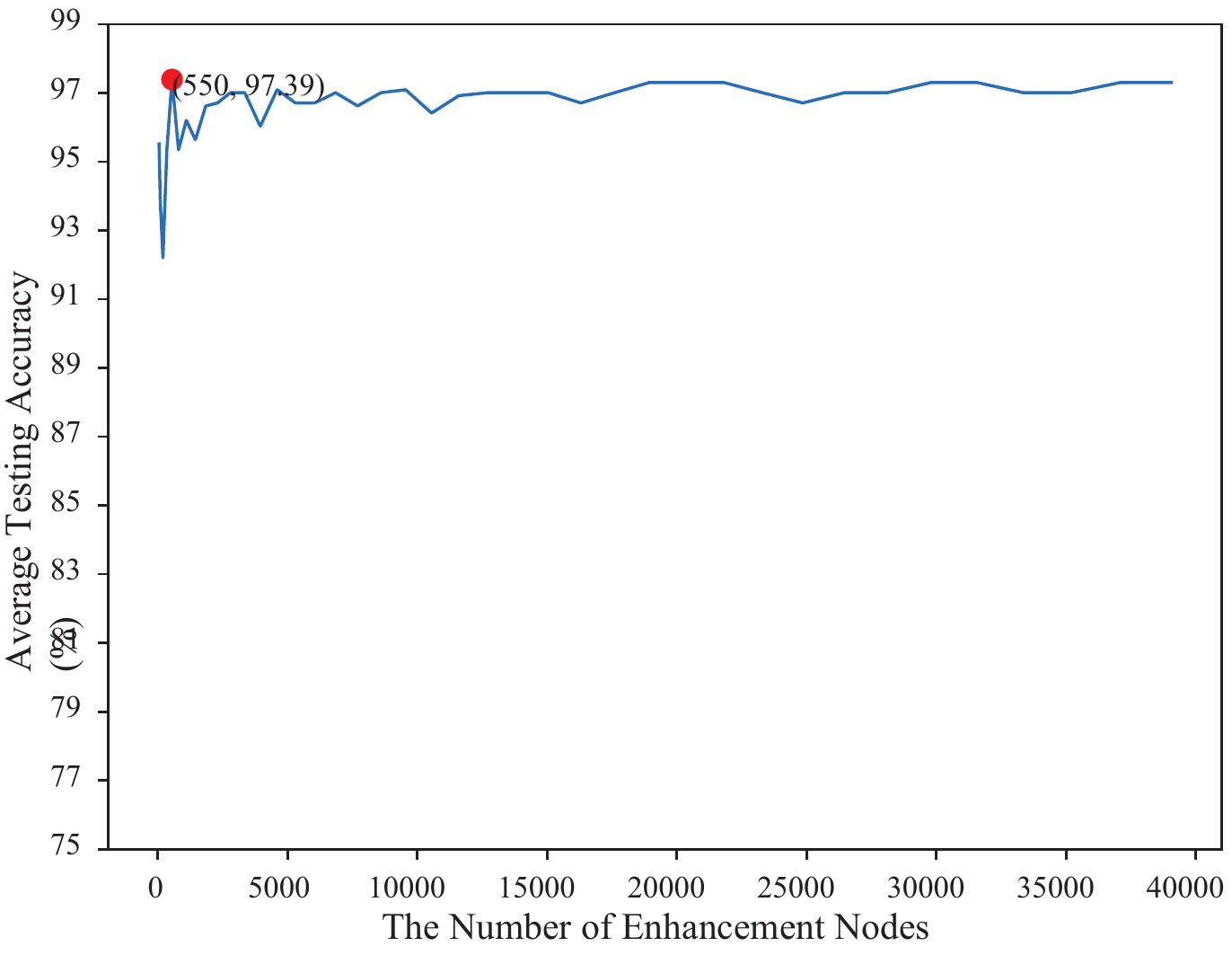}
  \caption{The effect of the number of enhancement nodes on the testing accuracy.}
  \label{fig:nodes}
\end{figure}

\textbf{Evaluation on The Number of Enhancement Nodes}\quad In DWnet, the number of enhancement nodes is a significant parameter which affects the accuracy of classifying. To make sure that our model can be reached the best performance, the relationship between testing accuracy and the number of enhancement nodes is shown in Fig. \ref{fig:nodes}. We initialize the number of enhancement nodes to 50 and add a multiple of 50 for each iteration. When the number of enhancement nodes is between 0 and 550, the overall trend of the curve first decreases and then increases. It tends to be stable until the number of enhancement nodes reaches around 500. And the peak appears when the number of enhancement nodes is around 550. As can be seen from Fig. \ref{fig:nodes}, when the number of enhancement nodes is 550, the average testing accuracy is the highest. The testing accuracy is improved from 200 to 550. And as the number of enhancement nodes increases, the testing accuracy is unstable. Furthermore, the more enhancement nodes DWnet has, the more time it takes to compute. Therefore, we set the number of enhanced nodes to 550 in DWnet.

\subsection{Evaluation on NTU RGB$+$D dataset}

\textbf{Evaluation on The NTU RGB$+$D Dataset}\quad To verify the generality of the proposed model, we evaluate the DWnet on the NTU RGB$+$D dataset that is a large scale dataset for human 3D human activity analysis \cite{NTU_RGB}. Due to the recognizing accuracy of BLS and HCNBLS on SBU Kinect dataset is much more dissatisfied than that of the HCN. We only compare the DWnet with the HCN on NTU RGB$+$D dataset, and the results are shown in Table \ref{tab:NTU}. In the CV and CS, the recognizing accuracy of DWnet is only 0.06\%, 0.23\% lower than that of the HCN, respectively. It is very close to the state-of-the-art performance. There are 60 kinds of actions in the NTU RGB$+$D dataset. Compared with the SBU Kinect dataset, actions in the NTU RGB$+$D dataset are more complicated. The dimension of the feature extracted by the PruHCN is higher than that of the SBU Kinect dataset. So it is a little labored for BLS to process the intricate features. Although the accuracy of recognition has a little decline, it can be seen from TABLE \ref{tab:timetable1} that the testing speed of DWnet is almost 100 times faster than that of HCN.

\begin{table}
  \renewcommand\arraystretch{1.7}
  \caption{Testing accuracy and testing time on the NTU RGB+D dataset. The unit of time is seconds/one sample.}
  \label{tab:NTU}
  \begin{tabular}{ccccc}
    \toprule
    Method & CV & \tabincell{c}{Testing \\ Time} & CS & \tabincell{c}{Testing \\ Time}\\
    \midrule
     HCN   & 89.90\% & 1.1e-2     & 84.30 \% & 1.2e-2  \\
     DWnet & 89.84\% & 2.26e-4  & 84.07 \% & 1.66e-4 \\
    \bottomrule
  \end{tabular}
\end{table}

\begin{figure}[htbp]
  \centering
  \includegraphics[scale=0.4]{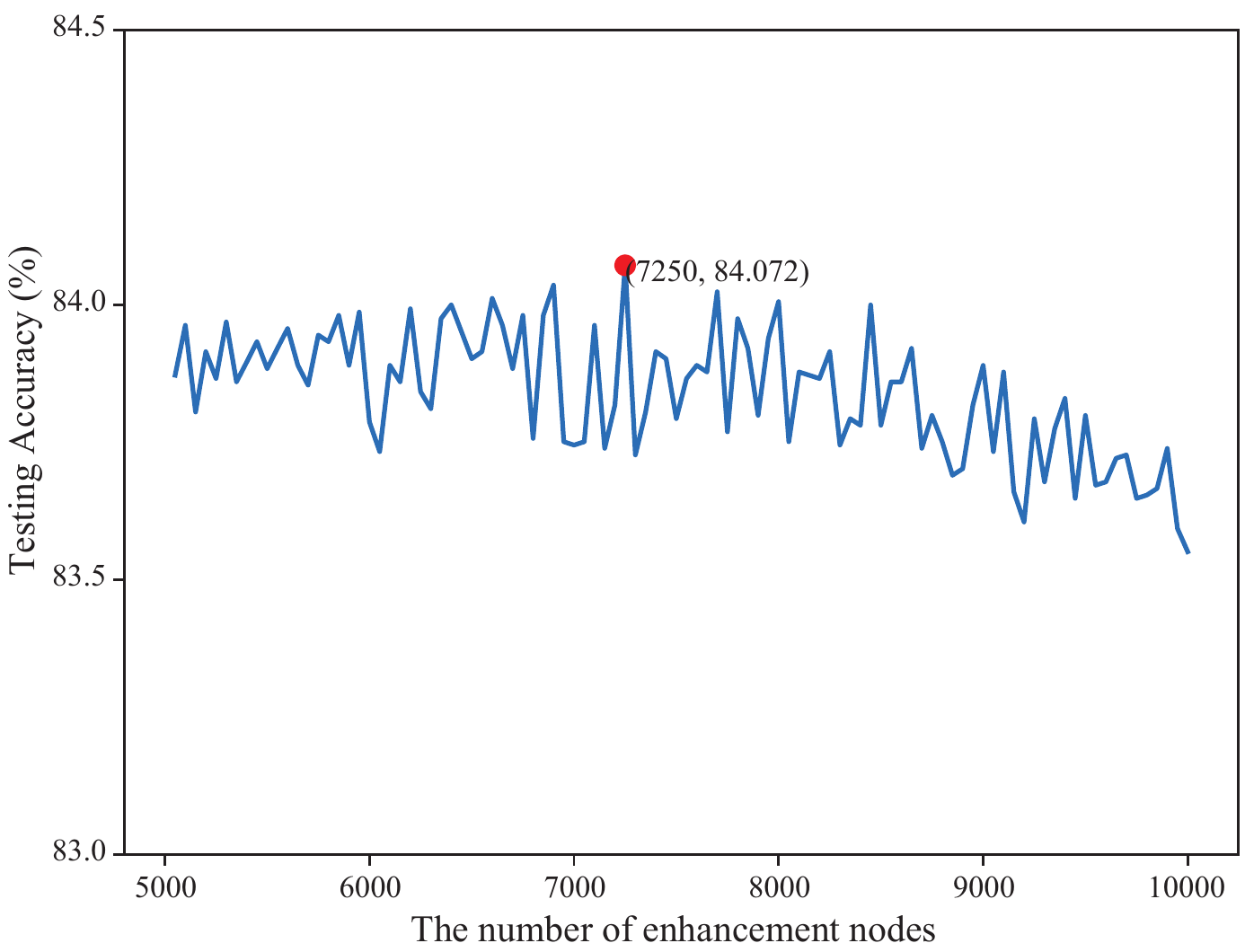}
  \caption{The effect of the number of enhancement nodes on the testing accuracy in the cross-subjects (CS).}
  \label{fig:CS_NTU}
\end{figure}
\begin{figure}[htbp]
  \centering
  \includegraphics[scale=0.4]{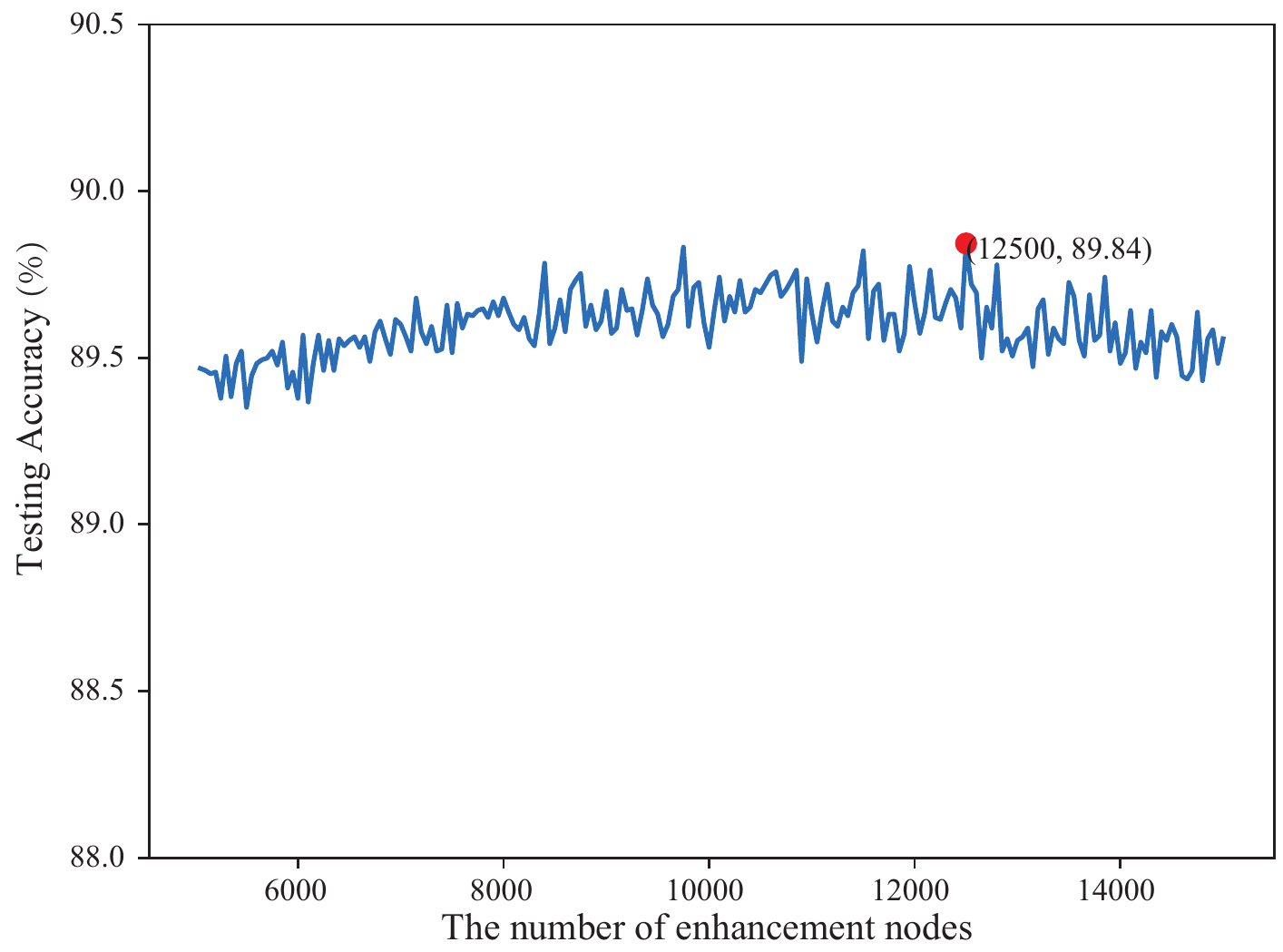}
  \caption{The effect of the number of enhancement nodes on the testing accuracy in the cross-views (CV).}
  \label{fig:CV_NTU}
\end{figure}

\textbf{Evaluation on The Number of Enhancement Nodes}\quad We have studied the relationship between the number of enhancement nodes and testing accuracy in CS, CV on the NTU RGB$+$D, separately, as shown in Fig. \ref{fig:CS_NTU} and Fig. \ref{fig:CV_NTU}. Due to the complexity of features is high, the number of enhancement nodes is initialized to 5000 to keep the approximation ability of the model. In CS and CV, the number of enhanced nodes increases from 5000 to 10000 and 14000 respectively, each time increasing by 50. The overall trend of the curves in Fig. \ref{fig:CS_NTU} and Fig. \ref{fig:CV_NTU} increases first and then decreases. In CS, the best result is obtained when the number of enhancement nodes is 7500. And in CV, the highest recognizing accuracy is reached when the number of enhancement nodes is 12500.

\section{Conclusion}
In this work, we have proposed a novel framework which combines deep learning structure with the broad learning system and introduced this model to recognize actions using 3D skeletal data. The DWnet combines the advantage of deep learning and broad learning. On one hand, DWnet can capture effective local and global spatial-temporal co-occurrence features by the deep structure. On the other hand, using broad learning system can expand features extracted by deep structure to higher dimensions so that sufficient local and global information can be obtained. In general, compared with the deep structure, DWnet saves lots of testing time and almost achieves real-time testing. Furthermore, the DWnet reduces the randomness during feature mapping and can capture better features than broad learning system can. Experimental results show that proposed DWnet model can achieve the state-of-the-art performance on the dataset. However, the DWnet is non-end-to-end model, so we will try to explore an end-to-end structure to recognize the action in future.


%

\appendices
\section*{Acknowledgment}
This work was supported partly by the National Natural Science Foundation of China Grant No. 61673192, No. U1613212, No. 61573219, the Fund for Outstanding Youth of Shandong Provincial High School ZR2016JL023, Shandong Provincial Key Research and Development Plan 2017CXGC1504, Dominant Discipline and Talent Team of Shandong Province Higher Education Institutions and the Basic Scientific Research Project of Beijing University of Posts and Telecommunications 2018RC31.

\ifCLASSOPTIONcaptionsoff
  \newpage
\fi



\bibliographystyle{IEEEtran}
\bibliography{IEEEabrv,reference}
%

\end{document}